\documentclass{article}
\usepackage{amsmath}
\usepackage{algorithm}
\usepackage[noend]{algpseudocode}
\usepackage{wrapfig}
\usepackage{arxiv}
\usepackage{graphicx}
\usepackage{enumitem}
\usepackage{amsmath}
\usepackage[utf8]{inputenc} 
\usepackage[T1]{fontenc}    
\usepackage[hidelinks]{hyperref}
\usepackage{url}            
\usepackage{booktabs}       
\usepackage{amsfonts}       
\usepackage{nicefrac}       
\usepackage{microtype}      
\usepackage{lipsum}
\usepackage{booktabs}
\usepackage{longtable}

\usepackage{array}
\newcolumntype{C}[1]{>{\centering\arraybackslash}p{#1}}

\usepackage{mathtools}
\usepackage[width=.9\textwidth]{caption}
\DeclarePairedDelimiterX{\infdivx}[2]{(}{)}{%
  #1\;\delimsize\|\;#2%
}

\newcommand{\infdiv}{\infdivx}

\usepackage[style=authoryear,sorting=nyt,citestyle=authoryearbrak,uniquelist=false,maxbibnames=99, natbib]{biblatex}

\usepackage{xcolor}

\title{Soft Actor-Critic for Discrete Action Settings}

\author{%
  Petros Christodoulou \\
  Imperial College London \\
  \texttt{petros.christodoulou18@imperial.ac.uk}
}

\begin{document}

\maketitle

\begin{abstract}

Soft Actor-Critic is a state-of-the-art reinforcement learning algorithm for continuous action settings that is not applicable to discrete action settings. Many important settings involve discrete actions, however, and so here we derive an alternative version of the Soft Actor-Critic algorithm that \textit{is} applicable to discrete action settings. We then show that, even without any hyperparameter tuning, it is competitive with the tuned model-free state-of-the-art on a selection of games from the Atari suite.
\end{abstract}

\section{Introduction}

Reinforcement Learning (RL) has famously made great progress in recent years, successfully being applied to settings such as board games \citep{Go}, video games \citep{DQN_Atari} and robot tasks \citep{OpenAI2018LearningDI}.  However, widespread adoption of RL in real-world domains has remained slow  primarily because of its poor sample efficiency which \citet{AKTR} see as a ”dominant concern in RL”.

\citet{SAC} provide the Soft Actor-Critic (SAC) algorithm which helps deal with this concern in continuous action settings. It has achieved model-free state-of-the-art sample efficiency in multiple challenging continuous control domains. Many domains however involve \textit{discrete} rather than continuous actions and in these environments SAC is not currently applicable. This paper derives a version of SAC that is applicable to discrete action domains and then shows that it is competitive with the model-free state-of-the-art for discrete action domains in terms of sample efficiency on a selection of games from the Atari \citep{Atari} suite. 

We proceed as follows: first we explain the derivation of Soft Actor-Critic for continuous action settings found in \citet{SAC} and \citet{SACapplications}, then we derive and explain the changes required to create a discrete action version of the algorithm, and finally we test the discrete action algorithm on the Atari suite. 

\section{Soft Actor-Critic}

Soft Actor-Critic \citep{SAC} attempts to find a policy that maximises the maximum entropy objective:

\begin{equation}
\pi^* = \underset{\pi}{\mathrm{argmax}} \sum_{t=0}^{T} E_{(s_t, a_t) \sim \tau_{\pi}}[\gamma^t (r(s_t, a_t) + \alpha \mathcal{H}(\pi(.|s_t))]
\end{equation}

where $\pi$ is a policy, $\pi^*$ is the optimal policy, $T$ is the number of timesteps, $r : S \times A \to \mathbb{R}$ is the reward function, $\gamma \in [0, 1] $ is the discount rate, $s_t \in S$ is the state at timestep t, $a_t \in A$ is the action at timestep t, $\tau_{\pi}$ is the distribution of trajectories induced by policy $\pi$, $\alpha$ determines the relative importance of the entropy term versus the reward and is called the temperature parameter, and $\mathcal{H}(\pi(.|s_t)$ is the entropy of the policy $\pi$ at state $s_t$ and is calculated as $\mathcal{H}(\pi(.|s_t)) = -\log \pi(.|s_t) $. 

To maximise the objective the authors use soft policy iteration which is a method of alternating between policy evaluation and policy improvement within the maximum entropy framework. 

The policy evaluation step involves computing the value of policy $\pi$. To do this they first define the soft state value function as:

\begin{equation}
V(s_t) \coloneqq E_{a_t \sim \pi}[Q(s_t, a_t) - \alpha \log(\pi(a_t|s_t))]
\label{eq:soft_state_value}
\end{equation}

They then prove that in a tabular setting (i.e. when the state space is discrete) we can obtain the soft q-function by starting from a randomly initialised function $Q : S \times A \to \mathbb{R}$ and repeatedly applying the modified Bellman backup operator $T^{\pi}$ given by: 

\begin{equation}
T^{\pi}Q(s_t, a_t) \coloneqq r(s_t, a_t) + \gamma E_{s_{t+1} \sim p(s_t, a_t)} [V(s_{t+1})]
\label{eq:sac_backup}
\end{equation}

where $p: S \times A \to S$ gives the distribution over the next state given the current state and action.

In the continuous state (instead of tabular) setting they explain that we instead firstly parameterise the soft q-function $Q_{\theta}(s_t, a_t)$ using a neural network with parameters $\theta$. Then we train the soft Q-function to minimise the soft Bellman residual:

\begin{equation}
J_{Q}(\theta) = E_{(s_t, a_t) \sim D}[\frac{1}{2}(Q_{\theta}(s_t, a_t)  - (r(s_t, a_t) + \gamma E_{s_{t+1}\sim p(s_t, a_t)} [V_{\bar{\theta}}(s_{t+1})]))^2]
\label{eq:soft_bellman_residual}
\end{equation}

where D is a replay buffer of past experiences and $V_{\bar{\theta}}(s_{t+1})$ is estimated using a target network for $Q$ and a monte-carlo estimate of \eqref{eq:soft_state_value} after sampling experiences from the replay buffer .

The policy improvement step then involves updating the policy in a direction that maximises the rewards it will achieve. To do this they use the soft Q-function calculated in the policy evaluation step to guide changes to the policy. Specifically, they update the policy towards the exponential of the new soft Q-function. Because they also want the policy to be tractable however, they restrict the possible policies to a parameterised family of distributions (e.g. Gaussian). To account for this, after updating the policy towards the exponential of the soft Q-function they then project it back into the space of acceptable policies using the information projection defined in terms of Kullback-Leibler divergence. So overall the policy improvement step is given by:

\begin{equation}
\pi_{\text{new}} = \underset{\pi \in \Pi}{\mathrm{argmin}} D_{\text{KL}}\infdiv[\bigg]{\pi(.|s_t)} {  \frac{\exp(\frac{1}{\alpha}Q^{\pi_{old}}(s_t,.))}{Z^{\pi_{\text{old}}}(s_t)}}
\label{eq:sac_policy_improvement}
\end{equation}

They note that partition function $Z^{\pi_{\text{old}}}(s_t)$ is intractable but does not contribute to the gradient with respect to the new policy and so it can be ignored. 

In the continuous state setting they parameterise the policy $\pi_{\phi}(a_t|s_t)$ using a neural network with parameters $\phi$ that outputs a mean and covariance that is then used to define a Gaussian policy. The policy parameters are then learned by minimizing the expected KL-divergence \eqref{eq:sac_policy_improvement} after multiplying by the temperature parameter $\alpha$ and ignoring the partition function $Z^{\pi_{\text{old}}}(s_t)$ as it does not impact the gradient:

\begin{equation}
J_{\pi}(\phi) = E_{s_t \sim D}[E_{a_t \sim \pi_{\phi}}[\alpha \log(\pi_{\phi}(a_t|s_t)) - Q_{\theta}(s_t, a_t)]]
\label{eq:policy_objective_before_reparam}
\end{equation}

This involves taking an expectation over the policy's output distribution which means errors cannot be backpropagated in the normal way. To deal with this they use the reparameterisation trick [\cite{reparatrick}] - instead of using the output of the policy network to form a stochastic action distribution directly, they combine its output with an input noise vector sampled from a spherical Gaussian. For example, in the one-dimensional case our network outputs a mean $m$ and standard deviation $s$. We could randomly sample our action directly $ a \sim N(m, s)$ but then we could not backpropagate the errors through this operation. So instead we do $a = m + s \epsilon $ where $\epsilon \sim N(0, 1)$ which allows us to backpropagate as normal. To signify that they are reparameterising the policy in this way they write:

\begin{equation}
a_t = f_{\phi}(\epsilon_t; s_t)
\label{eq:reparam}
\end{equation}

where $\epsilon_t \sim N(0, I)$. The new policy objective then becomes:

\begin{equation}
J_{\pi}(\phi) = E_{s_t \sim D, \epsilon_t \sim N}[\alpha \log(\pi_{\phi}(f_{\phi}(\epsilon_t; s_t)|s_t)) - Q_{\theta}(s_t, f_{\phi}(\epsilon_t; s_t))]
\label{eq:policy_objective_after_reparam}
\end{equation}

where $\pi_{\phi}$ is now defined implicitly in terms of $f_{\phi}$. They then go on to prove that in the tabular setting, alternating between policy evaluation and policy improvement as above will converge to the optimal policy. 

\citet{SACapplications} also provide an optional way of learning the temperature parameter so that we do not need to set it as a hyperparameter. They provide a long derivation for the temperature objective value, however because the details are not strictly relevant for our derivation of the discrete action version of SAC we do not repeat it here. The final objective they get to for the temperature parameter is however relevant and given by:

\begin{equation}
J(\alpha) = E_{a_t \sim \pi_t}[-\alpha (\log \pi_t(a_t|s_t) + \bar{H})]
\label{eq:temperature_objective}
\end{equation}

where $\bar{H}$ is a constant vector equal to the hyperparameter representing the target entropy.  They are unable to minimise this expression directly because of the expectation operator and so instead they minimise a monte-carlo estimate of it after sampling experiences from the replay buffer. 

Lastly, in practice the authors maintain two separately trained soft Q-networks and then use the minimum of their two outputs to be the soft Q-network output in the above objectives. They do this because \citet{TD3} showed that it helps combat state-value overestimation.

\section{Soft Actor-Critic for Discrete Action Settings (SAC-Discrete)}

We now derive a discrete action version of the above SAC algorithm. The first thing to note is that all the critical steps involved in deriving the objectives above hold whether the actions are continuous or discrete. All that changes is that $\pi_{\phi}(a_t|s_t)$ now outputs a probability instead of a density. Therefore the three objective functions $J_Q(\theta)$ \eqref{eq:soft_bellman_residual}, $J_{\pi}(\phi)$ \eqref{eq:policy_objective_before_reparam} and $J(\alpha)$ \eqref{eq:temperature_objective} still hold. We must however make five important changes to the process of optimising these objective functions:

i) It is now more efficient to have the soft Q-function output the Q-value of each possible action rather than simply the action provided as an input, i.e. our Q function moves from $Q: S \times A \to \mathbb{R}$ to $Q: S \to \mathbb{R}^{|A|}$. This was not possible before when there were infinitely many possible actions we could take. 

ii) There is now no need for our policy to output the mean and covariance of our action distribution, instead it can directly output our action distribution. The policy therefore changes from $\pi: S \to \mathbb{R}^{2|A|}$ to $\pi: S \to [0, 1]^{|A|}$ where now we are applying a softmax function in the final layer of the policy to ensure it outputs a valid probability distribution.

iii) Before, in order to minimise the soft Q-function cost $J_Q({\theta})$ \eqref{eq:soft_bellman_residual} we had to plug in our sampled actions from the replay buffer to form a monte-carlo estimate of the soft state-value function \eqref{eq:soft_state_value}. This was because estimating the soft state-value function in \eqref{eq:soft_state_value} involved taking an expectation over the action distribution.  However, now, because our action set is discrete we can fully recover the action distribution and so there is no need to form a monte-carlo estimate and instead we can calculate the expectation directly. This change should reduce the variance involved in our estimate of the objective $J_Q({\theta})$ \eqref{eq:soft_bellman_residual}. This means that we change our soft state-value calculation equation from \eqref{eq:soft_state_value} to:

\begin{equation}
 V(s_t) \coloneqq  \pi(s_t)^T[Q(s_t) - \alpha \log(\pi(s_t))]
\label{eq:new_soft_state_value}
\end{equation}

iv) Similarly, we can make the same change to our calculation of the temperature loss to also reduce the variance of that estimate. The temperature objective changes from \eqref{eq:temperature_objective} to:

\begin{equation}
J(\alpha) =  \pi_t(s_t)^T[-\alpha (\log (\pi_t(s_t)) + \bar{H})]
\label{eq:temperature_objective_new}
\end{equation}

v) Before, to minimise $J_{\pi}(\phi)$ \eqref{eq:policy_objective_before_reparam} we had to use the reparameterisation trick to allow gradients to pass through the expectations operator. However, now our policy outputs the exact action distribution we are able to calculcate the expectation directly. Therefore there is no need for the reparameterisation trick and the new objective for the policy changes from \eqref{eq:policy_objective_after_reparam} to:

\begin{equation}
J_{\pi}(\phi) = E_{s_t \sim D}[\pi_t(s_t)^T[\alpha \log(\pi_{\phi}(s_t)) - Q_{\theta}(s_t)]]
\label{eq:policy_objective_new}
\end{equation}

Combining all these changes, our algorithm for SAC with discrete actions (SAC-Discrete) is given by Algorithm \ref{alg:soft_actor_critic}.

\begin{algorithm}
\caption{Soft Actor-Critic with Discrete Actions (SAC-Discrete)}
\label{alg:soft_actor_critic}
\begin{algorithmic}
\State Initialise $Q_{\theta_1}: S \to \mathbb{R}^{|A|}$ , $Q_{\theta_2}: S \to \mathbb{R}^{|A|}$, $\pi_{\phi}: S \to [0, 1]^{|A|}$ \Comment{Initialise local networks}
\State Initialise $\bar Q_{\theta_1}: S \to \mathbb{R}^{|A|} , \bar Q_{\theta_2}: S \to \mathbb{R}^{|A|}$ \Comment{Initialise target networks}
\State $\bar \theta_1 \leftarrow \theta_1$, $\bar \theta_2 \leftarrow \theta_2$ \Comment{Equalise target and local network weights}
\State $\mathcal{D}\leftarrow\emptyset$ \Comment{Initialize an empty replay buffer}
\For{each iteration}
	\For{each environment step}
	    \State $a_t \sim \pi_{\phi}(a_t|s_t)$ \Comment{Sample action from the policy}
	    \State $s_{t+1} \sim p(s_{t+1}|s_t, a_t)$ \Comment{Sample transition from the environment}
	    \State $\mathcal{D} \leftarrow \mathcal{D} \cup \left\{(s_t, a_t, r(s_t, a_t), s_{t+1})\right\}$ \Comment{Store the transition in the replay buffer}
	\EndFor
	\For{each gradient step}
	    \State $\theta_i \leftarrow \theta_i - \lambda_Q \hat \nabla_{\theta_i} J(\theta_i)$ for $i \in \{1, 2\}$ \Comment{Update the Q-function parameters}

	    \State $\phi \leftarrow \phi - \lambda_\pi \hat \nabla_\phi J_\pi(\phi)$\Comment{Update policy weights}
	    \State $\alpha \leftarrow \alpha - \lambda \hat \nabla_\alpha J(\alpha)$ \Comment{Update temperature}
	    \State $\bar Q_i \leftarrow \tau Q_i + (1-\tau)\bar Q_i$ for $i\in\{1,2\}$\Comment{Update target network weights}
	\EndFor
\EndFor
  \textbf{Output} 
 $\theta_1$, $\theta_2$, $\phi $\Comment{Optimized parameters}
\end{algorithmic}
\end{algorithm}

\begin{figure}[h]
\centering
\includegraphics[width=0.8\textwidth]{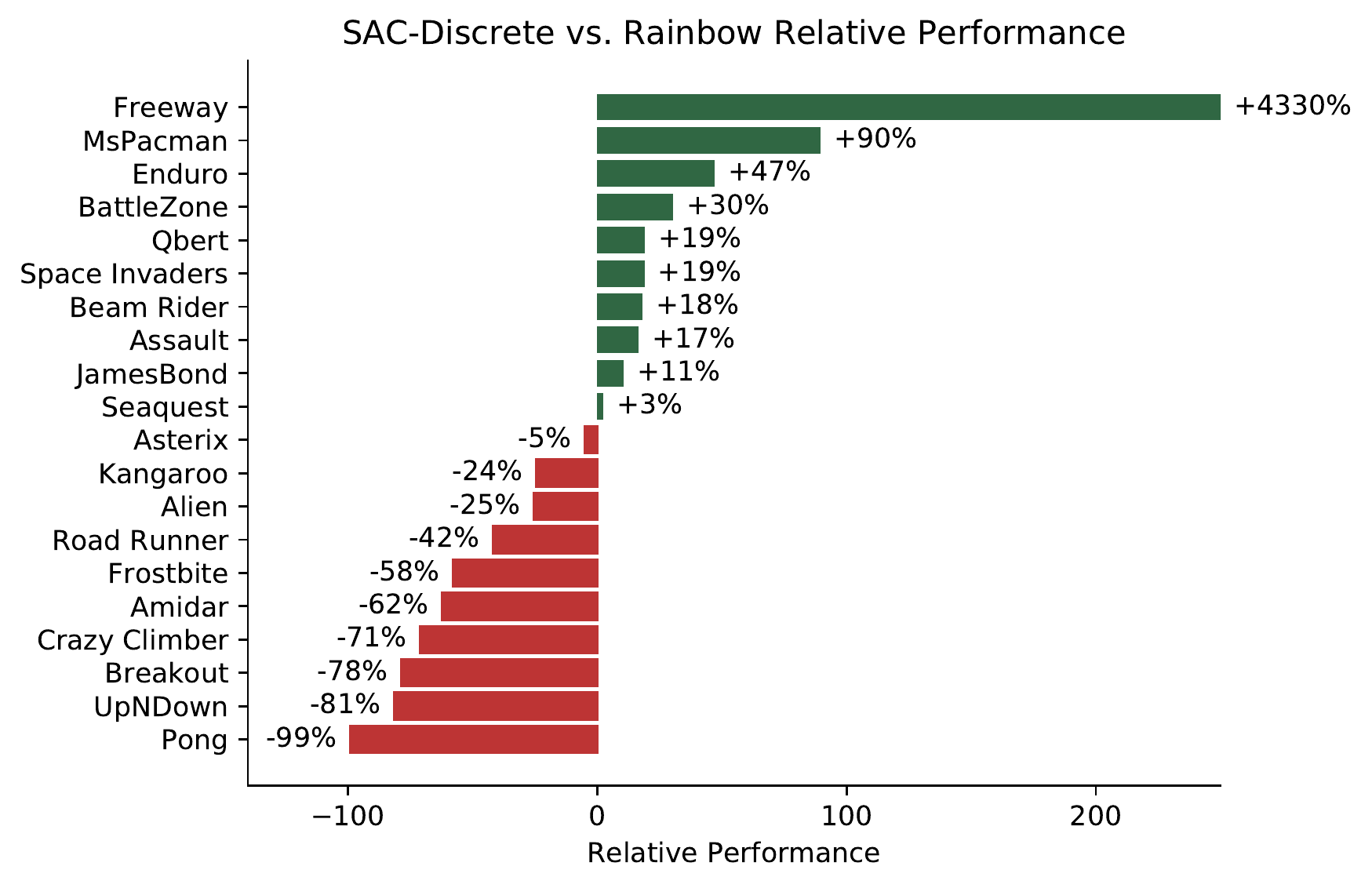}
\caption{\label{fig:sac_vs_rainbow} Comparing SAC-Discrete to Rainbow for 20 Atari games. The graph shows the average relative performance of SAC-Discrete over Rainbow over 5 random seeds where evaluation scores are calculated at the end of 100,000 steps of training. Note that no hyperparameter tuning was done to support the SAC scores compared to the Rainbow scores which benefited from substantial hyperparameter tuning}
\end{figure}
\section{Results}

To test the effectiveness of SAC-Discrete we run it for 100,000 steps on 20 Atari games for 5 random seeds each and compare its results with Rainbow which is a state-of-the-art model-free algorithm in terms of sample efficiency.  The games vary significantly and were chosen a piori and so we believe the results on these 20 games are a good estimate for relative performance on the whole Atari suite of 49 games.  We chose to run the algorithm for 100,000 steps because we are most interested in sample efficiency and \citet{model-Atari} demonstrated that Rainbow can make significant progress on Atari games within 100,000 steps.

For SAC-Discrete actions we did no hyperparameter tuning and instead used a mixture of the hyperparameters found in \citet{SACapplications} and \citet{model-Atari}. The hyperparameters can be found in Appendix \ref{sachyperparameters}. The Rainbow results we compare to come from \citet{model-Atari} and as they explain were the result of a significant amount of hyperparameter tuning.\footnote{For two games (Enduro and SpaceInvaders) \citet{model-Atari} provide no results for Rainbow and so for these two games only we ran the Rainbow algorithm ourselves. We used the Dopamine \parencite{dopamine} codebase to do this (as \citet{model-Atari} also did) along with the same (tuned) hyperaprameters used in \citet{model-Atari}. We share the code used to do this in the colaboratory notebook: https://colab.research.google.com/drive/11prfRfM5qrMsfUXV6cGY868HtwDphxaF} Therefore we are comparing the tuned Rainbow algorithm to our untuned SAC algorithm and so it is highly likely the relative performance of SAC could be improved if we spent time tuning its hyperparameters. 

We find that SAC-Discrete achieves a better score than Rainbow in 10 out of 20 games with a median performance of -1\%, maximum performance of +4330\% and minimum of -99\% - Figure \ref{fig:sac_vs_rainbow} summarises the results and Appendix \ref{sac_results} provides them in a table. Overall, we therefore consider the SAC-Discrete algorithm as roughly competitive with the model-free state-of-the-art on the Atari suite in terms of sample efficiency. 

\section{Conclusion}

The original Soft Actor-Critic algorithm achieved state-of-the-art results on numerous continuous action settings but was not applicable to discrete action settings. To correct this we have derived a version of the algorithm called SAC-Discrete that is applicable to discrete action settings and have shown that it performs competitively with the model-free state-of-the-art on the Atari suite even without any hyperparameter tuning. We provide a Python implementation of the algorithm at the project's GitHub repository.\footnote{https://github.com/p-christ/Deep-Reinforcement-Learning-Algorithms-with-PyTorch}   

\printbibliography

@article{OpenAI2018LearningDI,
  title={Learning Dexterous In-Hand Manipulation},
  author={OpenAI and Marcin Andrychowicz and Bowen Baker and Maciek Chociej and Rafal J{\'o}zefowicz and Bob McGrew and Jakub W. Pachocki and Arthur Petron and Matthias Plappert and Glenn Powell and Alex Ray and Jonas Schneider and Szymon Sidor and Josh Tobin and Peter Welinder and Lilian Weng and Wojciech Zaremba},
  journal={ArXiv},
  year={2018},
  volume={abs/1808.00177}
}

@article{SACapplications,
author = {T. Haarnoja and A. Zhou and K. Hartikainen and G. Tucker and S. Ha, J. Tan and V. Kumar and H. Zhu and A. Gupta and P. Abbeel and S. Levine},
title = {Soft Actor-Critic Algorithms and Applications},
journal = {arXiv preprint},
year= {2019}
}

@article{AKTR,
title={Scalable Trust-Region Method for Deep Reinforcement
Learning Using Kronecker-Factored Approximation},
author={Y. Wu and E. Mansimov and S. Liao and R. Grosse and J. Ba},
journal={Advances in Neural Information Processing Systems},
year={2017}
}

@article{Atari,
title={The Arcade Learning Environment: An Evaluation Platform for General Agents}, 
author={M. Bellemare and Y. Naddaf and J. Veness and M. Bowling},
journal = {Journal of Artificial Intelligence},
year={2013}
}

@article{Go,
title={Mastering the Game of Go Without Human Knowledge},
author={D. Silver and J. Schrittwieser and K. Simonyan and I. Antonoglou and A. Huang and A. Guez and T. Hubert and L. Baker and M. Lai and A. Bolton and Y. Chen and T. Lillicrap and F. Hui and L. Sifre and G. Driessche and T. Graepel and D. Hassabis},
journal={Nature},
year={2017}}

@article{DQN_Atari,
title={Human-Level Control Through Deep Reinforcement Learning},
author={V. Mnih and K. Kavukcuoglu and D. Silver and A. Rusu and J. Veness and M. Bellemare and A. Graves and M. Riedmiller and A. Fidjeland and G. Ostrovski and S. Peterson and C. Beattie and A. Sadik and I. Antonoglou and H. King and D. Kumaran and D. Wierstra and S. Legg and D. Hassabis},
journal={Nature},
year={2015}}

@article{SAC,
  title={Soft Actor-Critic: Off-Policy Maximum Entropy Deep Reinforcement Learning with a Stochastic Actor},
  author={T. Haarnoja and A. Zhou and P. Abbeel and S. Levine},
  journal={International Conference on Learning Representations},
  year={2018}
}

@article{model-Atari,
author = {L. Kaiser and M. Babaeizadech and P. Milos and B. Osinski and R. Campbell and K. Czechowski and D. Erhan and C. Finn and P. Kozakowski and S. Levine and A. Mohiuddin and R. Sepassi and G. Tucker and H. Michalewski},
title = {Model Based Reinforcement Learning for Atari},
journal = {arXiv preprint},
year= {2019}
}

@article{reparatrick,
author = {D. Kingma and M. Welling},
title = {Auto-Encoding Variational Bayes},
journal = {International Conference on Learning Representations},
year= {2013}
}

@article{rainbow,
author = {M. Hessel and J. Modayil and H. van Hasselt and T. Schaul and G. Ostrovski and W. Dabney and D. Horgan and B. Piot and M. Azar and D. Silver},
title = {Rainbow: Combining Improvements in Deep Reinforcement Learning},
journal = {arXiv preprint},
year= {2017}
}

@article{dopamine,
author = {P. Castro and S. Moitra and C. Gelanda and S. Kumar and M. Bellemare},
title = {Dopamine: A Research Framework for Deep Reinforcement Learning},
journal = {arXiv preprint},
year= {2018}
}

@article{TD3,
author = {S. Fujimoto and H. van Hoof and D. Meger},
title = {Addressing Function Approximation Error in Actor-Critic Methods},
journal = {arXiv preprint},
year= {2018}
}
\clearpage

\section*{Appendix}
\appendix

\section{SAC and Rainbow Atari Results}
\label{sac_results}

\begin{table}[h]
\centering

\bgroup
\def\arraystretch{1.1}
\caption{SAC and Rainbow results on 20 Atari games. The mean SAC result of 5 random seeds is shown with the standard deviation in brackets. As a benchmark we also provide a column indicating the score an agent would get if it acted purely randomly. The Rainbow results come from \cite{model-Atari}.}

\begin{tabular}{|c|c|c|c|}
\hline
\textbf{Game}  & \textbf{Random} & \textbf{Rainbow} & \textbf{SAC}     \\ \hline

Freeway        & 0.0             & 0.1              &$\underset{(9.9)}{4.4}$             \\ \hline
MsPacman       & 235.2           & 364.3            &$\underset{(141.8)}{690.9}$      \\ \hline
Enduro         & 0.0             & 0.53              &$\underset{(0.8)}{0.8}$         \\ \hline
BattleZone          & 2895.0           & 3363.5            &$\underset{(1163.0)}{4386.7}$    \\ \hline
Qbert          & 166.1           & 235.6            &$\underset{(124.9)}{280.5}$        \\ \hline
Space Invaders & 148.0           & 135.1            &$\underset{(17.3)}{160.8}$         \\ \hline
Beam Rider     & 372.1           & 365.6            &$\underset{(44.0)}{432.1}$      \\ \hline
Assault          & 233.7           & 300.3            &$\underset{(40.0)}{350.0}$         \\ \hline
JamesBond           & 29.2           & 61.7            &$\underset{(26.2)}{68.3}$       \\ \hline
Seaquest       & 61.1            & 206.3            &$\underset{(59.1)}{211.6}$            \\ \hline
Asterix        & 248.8           & 285.7            &$\underset{(33.3)}{272.0}$      \\ \hline
Kangaroo        & 42.0             & 38.7              &$\underset{(55.1)}{29.3}$           \\ \hline
Alien          & 184.8           & 290.6            &$\underset{(43.0)}{216.9}$           \\ \hline
Road Runner    & 0.0             & 524.1            &$\underset{(557.4)}{305.3}$       \\ \hline
Frostbite      & 74.0            & 140.1            &$\underset{(16.3)}{59.4}$       \\ \hline
Amidar         & 11.8            & 20.8             &$\underset{(5.1)}{7.9}$            \\ \hline
Crazy Climber  & 7339.5          & 12558.3          &$\underset{(600.8)}{3668.7}$    \\ \hline
Breakout       & 0.9             & 3.3              &$\underset{(0.5)}{0.7}$              \\ \hline
UpNDown         & 488.4            & 1346.3             &$\underset{(176.5)}{250.7}$          \\ \hline
Pong           & -20.4           & -19.5            &$\underset{(0.0)}{-20.98}$          \\ \hline

\end{tabular}
\egroup

\end{table}

\clearpage
\section{SAC-Discrete Hyperparameters}
\label{sachyperparameters}

\begin{table}[h]
\centering

\bgroup
\def\arraystretch{1.1}
\caption{Hyperparameters used for SAC-Discrete results}
\label{SAChyperparms}

\begin{tabular}{|c|c|c|}
\hline
\textbf{Hyperparameter}  & \textbf{Value}  \\ \hline
Layers & 3 convolutional layers and 2 fully connected layers \\ \hline
Convolutional channels per layer & [32, 64, 64] \\ \hline
Convolutional kernel sizes per layer & [8, 4, 3] \\ \hline
Convolutional strides per layer & [4, 2, 1] \\ \hline
Convolutional padding per layer & [0, 0, 0] \\ \hline
Fully connected layer hidden units & [512, number of moves in game] \\ \hline
Batch size         & 64             \\ \hline
Replay buffer size         & 1,000,000            \\ \hline
Discount rate         & 0.99             \\ \hline
Steps per learning update         & 4            \\ \hline
Learning iterations per round         & 1             \\ \hline
Learning rate         & 0.0003            \\ \hline
Optimizer         & Adam             \\ \hline
Weight initialiser         & He            \\ \hline
Fixed network update frequency        & 8000            \\ \hline
Loss        & Mean squared error            \\ \hline
Clip rewards        & Clip to [-1, +1]             \\ \hline
Initial random steps & 20,000             \\ \hline
Entropy target & 0.98 * (-log (1 / |A|)) \\ \hline
\end{tabular}
\egroup
\end{table}

\end{document}